\documentclass[runningheads]{llncs}
\usepackage{graphicx}

\usepackage{tikz}
\usepackage{comment}
\usepackage{amsmath,amssymb} 
\usepackage{color}

\usepackage[accsupp]{axessibility}  

\usepackage[width=122mm,left=12mm,paperwidth=146mm,height=193mm,top=12mm,paperheight=217mm]{geometry}

\begin{document}
\pagestyle{headings}
\mainmatter

\title{Rethinking Query-Key Pairwise Interactions in Vision Transformers
} 

\author{Cheng Li
\quad
Yangxing Liu
\\
\{cheng16.li, yangxing.liu\}@tcl.com
}
\institute{TCL Research (Wuhan)}

\maketitle

\begin{abstract}
Vision Transformers have achieved state-of-the-art performance in many visual tasks. Due to the quadratic computational and memory complexities of self-attention, recent works either apply attention only to low-resolution inputs or restrict the receptive field to a small local region. 
To overcome these limitations, we propose key-only attention, which excludes query-key pairwise interactions and uses a compute-efficient saliency-gate to obtain attention weights, modeling local-global interactions in all stages. Key-only attention has linear computational and memory complexities w.r.t input size. We use alternate layout to hybridize convolution and attention layers instead of grafting which is suggested by previous works, so that all stages can benefit from both spatial attentions and convolutions.
We leverage these improvements to develop a new self-attention model family, LinGlos, which reach state-of-the-art accuracies on the parameter-limited setting of ImageNet classification benchmark, and outperform baselines significantly in downstream tasks, e.g., COCO object detection and ADE20K semantic segmentation.
\end{abstract}

\section{Introduction}
Vision Transformers\cite{ViT,Swin,DeiT,PVT,tnt} have attracted increasing interests since they have achieved state-of-the-art performance in many vision tasks. 
Unlike convolution kernel, which is an input-independent parameter of static value, attention weight dynamically depends on the representation of the input, which could lead to higher model capacity\cite{AAConv,HaloNet,CoAtNet}. Another advantage of attention is the ability to establish global dependencies\cite{CoAtNet}. 

Let
$X \in R^{N\times F}$denote a sequence of $N$ feature vectors of dimensions $F$ . The input $X$ is projected by 3 matrices $W_Q \in R^{F\times D}$,$W_K \in R^{F\times D}$,$W_V \in R^{F\times M}$,to representations $Q$ , $K$ and $V$. The output of attention can be formulated as:
\begin{equation}
Y=\text{softmax}(\frac{QK^T}{\sqrt{D}})V
\end{equation}

The computational and memory complexities of $QK^T$are quadratic in the number of visual elements, which may incur high latency and a lot of memory\cite{Lambdanetworks}.

To address this issue, recent works either restrict the receptive field of attention a small local region\cite{Swin,HaloNet} , or apply attention only to low-resolution feature maps through aggressive sub-sampling feature maps' resolution in early stages\cite{ViT,DeiT}, or using attention only in later stages\cite{CoAtNet}.These strategies deprive the ability of the network to model long-range interactions in its early and middle stages or produce multi-scale feature maps\cite{GSA}. Is computing query-key pairwise interactions $QK^T$really necessary?

In this work, we propose a network architecture by replacing dot-product attention with key-only attention, which obtains the attention weights of visual elements in all positions through a compute-efficient saliency-gate, summarizing the global context according to the attention weights, and modeling local-global interactions in all stages. The architecture enjoys linear computational complexity w.r.t. input's size and incurs much less latency and memory when processing high-resolution feature maps.
\begin{figure}
\centering
\includegraphics[height=5cm]{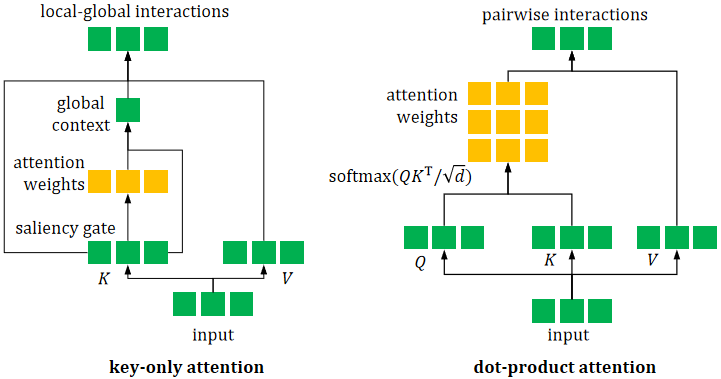}
\caption{The proposed key-only attention(left), which uses a compute-efficient saliency-gate to obtain attention weights, and excludes pairwise interactions.}
\label{fig:key_only_attention}
\end{figure}

To take advantages of both convolution and attention, previous works combine convolution and attention layers by grafting\cite{CoAtNet,HaloNet,Lambdanetworks,Non-local}: convolutions are in early stages and attentions in later stages. This is due to the computational inefficiency of spatial attention on high-resolution inputs\cite{Lambdanetworks,Non-local} and the prior that convolution is better at processing local patterns that are more common in early stages\cite{CoAtNet}. Different from these works, we use alternate layout: each stage includes both convolution and attention. We hypothesize that each stage's preference for convolution and attention may depend on specific visual tasks. Besides, several works\cite{PiT,ViT} show that long-range dependencies help in early stages.

We also propose residual parallel conditional positional encoding(RP-CPE), which is a modification of conditional positional encoding\cite{CPE} and yields further cost-free improvement. RP-CPE can be not only used in our architecture, but also easily and flexibly plugged into other vision transformers.

Leveraging these improvements, we develop a new self-attention visual bakbone family, LinGlos (\textbf{lin}ear-complexity and \textbf{glo}bal-local interaction). Compared with existing CNNs, vision transformers and hybrid models, every stage in LinGlos combines desirable properties in Table~\ref{table:properties}.
\setlength{\tabcolsep}{4pt}
\begin{table}
\begin{center}
\caption{Desirable properties in different architectures}
\label{table:properties}
\begin{tabular}{lccccc}
\hline\noalign{\smallskip}
Desirable properties & ResNet\cite{resnet} & Swin\cite{Swin} & ViT\cite{ViT} & CoAtNet\cite{CoAtNet} & LinGlo \\
\noalign{\smallskip}
\hline
\noalign{\smallskip}
Linear complexity  & {\checkmark} & {\checkmark} & {} & {} & {\checkmark}\\
Content-based interaction & {} & {\checkmark} & {\checkmark} & {\checkmark} & {\checkmark}\\
Global receptive field & {} & {} & {\checkmark} & {\checkmark} & {\checkmark}\\

\hline
\end{tabular}
\end{center}
\end{table}
\setlength{\tabcolsep}{1.4pt}

We evaluate the proposed LinGlos on several different tasks, including image classification, object detection and semantic segmentation .LinGlos reach state-of-the-art accuracies on the parameter-limited setting of the ImageNet classification benchmark, and outperform baselines significantly in downstream tasks, e.g., object detection and semantic segmentation.Code will be publicly available.
\section{Related Works }
\subsection{Vision Transformers}
ViT\cite{ViT} has set off a wave of replacing CNNs by transformers in computer vision. It uses global spatial dot-product attention whose computational complexity is quadratic w.r.t. spatial size.To avoid high latency and a lot of memory, ViT uses a aggressive sub-sampling strategy that shrinks the input's spatial size at the very beginning, which results that ViT can only produce low-resolution feature maps and is challenging to directly adapt it to pixel-level dense predictions.
Recent works\cite{PVT,Swin,Focal,PiT,Twins,CPE} use progressive sub-sampling strategies to produce multi-scale feature maps.To improve the computational efficiency on high-resolution inputs, PVT\cite{PVT} shrinks the spatial size of $K$ and $V$ before attention operations. Its computational complexity is still quadratic so that PVT is more suitable for tasks with low or medium-resolution inputs. Swin-transformer\cite{Swin} limits self-attention computation to non-overlapping local windows while also allowing for connections between adjacent windows, which may limit attention's ability to model long-range interactions.
In Focal transformer\cite{Focal} ,each token attends its closest surrounding tokens at fine granularity and the tokens far away at coarse granularity, which introduces extra computational and memory cost. 
\subsection{Spatial Attention without Pairwise Interactions}
Several works have questioned the necessity of pairwise interaction. Synthesizer\cite{Synthesizer} learns to synthesize the self-alignment matrix without token-token interactions and shows that token-token (query-key) interactions are not necessary to achieve good performance with transformer models on certain linguistic tasks .Empirical Attention\cite{EmpiricalAttention} ablates various spatial attention elements, including (1) the query and key content, (2) the query content and relative position, (3) the key content only, and (4) the relative position only.They find that the first two play a minor role in self-attention, and that proper combination of deformable\cite{deformable} convolution and the key content only term
is more compute-efficient in visual recognition tasks.Following \cite{EmpiricalAttention}, we use the terminology "key-only-attention" since we share the inherent spirit , while our formulations are quite different.
\subsection{Linear Computational Complexity Transformers}
\cite{Performer,LinearTransformer,linformer,fastformer,cosformer,Linear_transformers_2} propose transformers with linear computational complexity w.r.t. sequence length which perform competitively or better in tasks with long-context scenarios, such as protein sequence modeling, ASR, language modeling with long-sequence input. The design of LinGlos refers to these works. In our practice, directly applying linear complexity transformers to visual tasks performs poor . We find that the combination with some inductive bias from CNNs is crucial, such as hierarchical representations and hybridizing with convolutions.

\section{Method }

\subsection{Key-Only Attention}
Fig.~\ref{fig:key_only_attention} illustrates the key-only attention. We exclude the queries and no longer compute query-key pairwise interactions. Attention weights of visual elements are computed only from keys. The input $X \in R^{N\times F}$is projected by two matrices $W_K \in R^{F\times D}$,$W_V \in R^{F\times D}$ to corresponding representations $K$,$V \in R^{N\times D}$. A saliency-gate is used for determining the attention weights of keys: $K$ is projected by the matrix $W_{saliency} \in R^{D\times 1}$and normalized over all positions with softmax to produce attention map $A \in R^{N}$:
\begin{equation}
A=\text{softmax}(\frac{KW_{\text{saliency}}}{\sqrt{D}})
\end{equation}
The global context is computed as a weighted sum of keys according to the attention weights:
\begin{equation}
G=\sum\limits_{i=1}^N A_{i}K_{i}
\end{equation}
The output is calculated as :
\begin{equation}
Y=((G\odot V )U_1+K)U_2
\end{equation}
where $\odot$ denotes the element-wise product between the global key context and each value , building local-global interactions, and $U_1,U_2 \in R^{D\times D}$are linear projection matrices.

The attention weights in key-only attention differ from those in existing vision transformers\cite{HaloNet,Swin,ViT}, which measure the compatibility of visual element pairs, but bear some similarities to human visual attention\cite{HumanAttention} which allows for salient features to come to the forefront.

Key-only attention has linear computational complexity w.r.t. input's spatial size so that it can be applied in all stages, from the high-resolution input at the beginning to the low-resolution input at the end.

\subsection{Residual Parallel Conditional Positional Encoding}
\begin{figure}
\centering
\includegraphics[height=2.4cm]{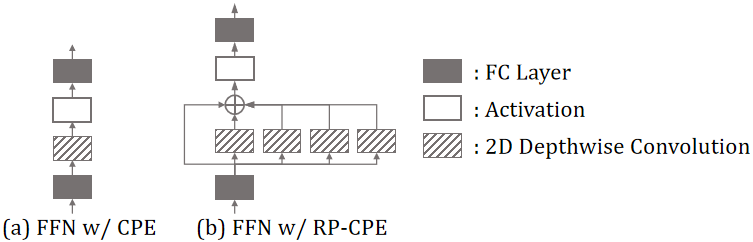}
\caption{The structures of CPE and the proposed RP-CPE }
\label{fig:rpcpe}
\end{figure}

Because convolutions can implicitly encode the position information\cite{Conv_Pos,Conv_Pos_2}, 
some works\cite{CPE,Twins} insert depth-wise convolution operations in FFN layers as illustrated in Fig.~\ref{fig:rpcpe} (a) , which called conditional positional encoding(CPE).
Based on CPE, we propose an improved version called residual parallel conditional positional encoding(RP-CPE). We 
simply add a skip connection and several other CPE branches in parallel as illustrated in Fig.~\ref{fig:rpcpe}(b) and re-parameterizing them into a single branch when testing and deploying the network. We use the re-parameterizing method proposed by \cite{DBB}.

\subsection{Overall Architecture}
\begin{figure}
\centering
\includegraphics[height=3.3cm]{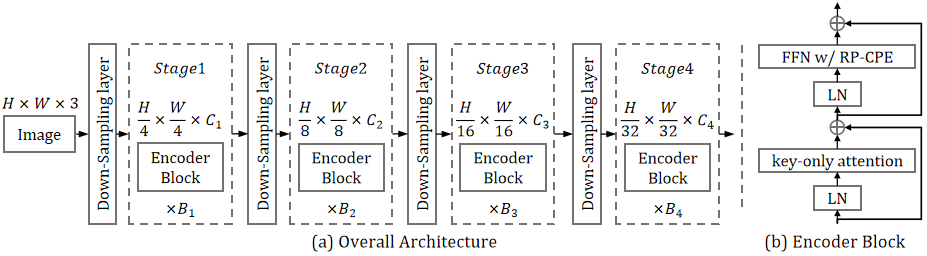}
\caption{Overview of LinGlos' architecture and compositions of encoder block.}
\label{fig:arg}
\end{figure}

An overview of LinGlos' architecture is presented in Fig.~\ref{fig:arg}(a). Following\cite{ConvNeXt,Swin}, we use separate down-sampling layers between stages to shrink spatial size and increase channel dimension. Down-sampling layers are implemented as regular convolution layers.
Layer normalizations\cite{layernorm} are added wherever spatial resolution is changed to stabilize training. Each encoder block is composed of a key-only attention layer, a FFN with RP-CPE, skip connections and layer normalizations as illustrated in Fig.~\ref{fig:arg}(b). 

\subsection{Architecture Details }

We construct different LinGlo variants, LinGlo-b0/b1/b2.
The parameter numbers of LinGlo-b1 and LinGlo-b2 are comparable to those of PVT-Tiny and PVT-Small.
These variants only differ in the number of channels C, and the number of blocks B in each stage. We summarize the configurations below: 

LinGlo-b0: C = (32, 64, 160, 256), B = (2, 3, 3, 2)

LinGlo-b1: C = (64, 128, 320, 512), B = (2, 3, 3, 2)

LinGlo-b2: C = (64, 128, 320, 512), B = (3, 5, 9, 3)

Attention head numbers are (1, 2, 5, 8), and expansion ratios of FFNs are (8, 8, 4, 4) as in PVT\cite{PVT}. The convolution kernel size, stride, padding size are (7, 4, 3) in the first down-sampling layer , and are (3, 2, 1) in other down-sampling layers. We use GELU\cite{gelu} as activation function. Each RP-CPE includes four 3$\times$3 depth-wise convolution branches.

\section{Experiments}

Experiments are conducted on ImageNet-1K image classification\cite{imagenet}, COCO object detection\cite{coco}, and ADE20K semantic segmentation\cite{ade20k} to evaluate the proposed method.

\subsection{Image classification}

\textbf{Setting.} ImageNet-1K dataset\cite{imagenet} consists of 1000 object classes with 1.2M training images and 50K validation images. We report ImageNet-1K top-1 accuracy on the validation set. During training, we follow DeiT\cite{DeiT} and apply random cropping, random horizontal flipping\cite{horizontal_flipping}, label-smoothing\cite{labelsmoothing}, Mixup\cite{mixup}, CutMix\cite{cutmix}, random erasing\cite{Random_erasing}, stochastic depth\cite{stochastic} and repeated augmentation\cite{DeiT} as data augmentations or regularizations.
We train LinGlos for 300 epochs from scratch using AdamW\cite{adamw} with a momentum of 0.9, a learning rate of 0.001, a weight decay of 0.05.There is a 5-epoch linear warm-up and a cosine decaying\cite{cosine} schedule afterward. The total batch size is 1024.To benchmark, we apply a center crop on the validation set, where a 224$\times$224 patch is cropped to evaluate the classification accuracy. 

\textbf{Result.} Table~\ref{table:imagenet_result} shows that LinGlos reach state-of-the-art accuracies on ImageNet-1K classification.
Compared to PVT\cite{PVT}, LinGlos have similar parameters and greatly improved accuracy.
LinGlo-b1 is 4.6\% higher than PVT-Tiny, and LinGlo-b2 is 2.9\% higher than PVT-Small. Compared to other recent counterparts, LinGlos also have advantages in terms of accuracies and parameters. LinGlo-b2 achieves 82.7\% ImageNet top-1 accuracy, which is 1.4\% higher than Swin Transformer-Tiny\cite{Swin} and 0.6\% higher than ConvNeXt-T\cite{CoAtNet}, while LinGlo-b2's parameters are fewer. 
\setlength{\tabcolsep}{4pt}
\begin{table}
\begin{center}
\caption{ImageNet-1K image classification result. All models are trained for 300 epochs from scratch with the input size of 224$\times$224.LinGlos have greatly improved accuracy.}
\label{table:imagenet_result}
\begin{tabular}{lcc}
\hline\noalign{\smallskip}
Method & Param(M) & Top-1 Acc(\%)\\
\noalign{\smallskip}
\hline
\noalign{\smallskip}
Deit-Tiny/16\cite{DeiT}  & {5.7} & {72.2} \\
LinGlo-b0(Ours)  & {3.4} & {\textbf{72.5}} \\
\hline
\noalign{\smallskip}
ResNet18\cite{resnet}  & {11.7} & {69.8} \\
LambdaNet-50\cite{Lambdanetworks}& {15.0} & {78.4} \\
PVT-Tiny\cite{PVT}  & {13.2} & {75.1} \\
LinGlo-b1(Ours)  & {13.1} & {\textbf{79.7}} \\
\hline
\noalign{\smallskip}
ResNet50\cite{resnet}  & {25.6} & {79.8} \\
DeiT-Small/16\cite{DeiT}  & {22.1} & {79.9} \\
TNT-S\cite{tnt}  & {23.8} & {81.3} \\
Swin-T\cite{Swin}  & {29.0} & {81.3} \\
Twins-SVT-S\cite{Twins}  & {24.0} & {81.3} \\
Focal-Tiny\cite{Focal}  & {29.1} & {82.2} \\
CoAtNet-0\cite{CoAtNet}  & {25} & {81.6} \\
ConvNeXt-T\cite{ConvNeXt}  & {29} & {82.1} \\
LinGlo-b2(Ours)  & {25.4} & {\textbf{82.7}} \\
\hline
\end{tabular}
\end{center}
\end{table}
\vspace{-0.5cm}
\setlength{\tabcolsep}{1.4pt}
\subsection{Object Detection}
\textbf{Setting.} All models are trained on COCO train2017\cite{coco} (118k images) and evaluated on val2017 (5k images). Following PVT\cite{PVT}, we verify the effectiveness of LinGlos' backbone on top of RetinaNet\cite{retinanet} detection head with FPN\cite{fpn} neck in mmdetection\cite{mmdetection}. We use the weights pre-trained on ImageNet-1K to initialize the backbone and Xavier\cite{Xavier} to initialize the newly added layers. All models are trained with a batch size of 16 and optimized by AdamW\cite{adamw} with an initial learning rate of 0.0001 for 12 epochs.The learning rate is multiplied by 0.1 after the 8-th and the 11-th epoch. The training image is resized to have a shorter side of 800 pixels, while the longer side does not exceed 1,333 pixels. In the testing phase, the shorter side of the input image is fixed to 800 pixels. 

\textbf{Result.} As reported in Table~\ref{table:coco_result}, LinGlos significantly outperform PVTs\cite{PVT} on top of RetinaNet\cite{retinanet} with similar model size. LinGlo-b1 is 4.8AP higher than PVT-Tiny, and LinGlo-b2 is 4.3AP higher than PVT-Small. Compared to recent counterparts, LinGlo-b2 achieves 44.7AP, which is 3.2AP higher than Swin Transformer-Tiny\cite{Swin} while LinGlo-b2's parameters are fewer. 
\setlength{\tabcolsep}{4pt}
\begin{table}
\begin{center}
\caption{COCO detection result.All models are trained for 12 epochs. LinGlos significantly outperform counterparts on top of RetinaNet\cite{retinanet} with similar model size.}
\label{table:coco_result}
\begin{tabular}{lccccccc}
\hline\noalign{\smallskip}
Backbone & Param(M) & AP & AP$_{50}$ & AP$_{75}$ & AP$_{S}$ & AP$_{M}$ & AP$_{L}$\\
\noalign{\smallskip}
\hline
\noalign{\smallskip}
LinGlo-b0(Ours)  & {13.0} & {38.0} & {58.3} & {40.2} & {22.8} & {41.2} & {49.6} \\
\noalign{\smallskip}
\hline
\noalign{\smallskip}
{ResNet18}\cite{resnet} & {21.3}  & {31.8} & {49.6} & {33.6} & {16.3}  & {34.3} & {43.2}\\
{PVT-Tiny}\cite{PVT} & {23.0}  & {36.7} & {56.9} & {38.9} & {22.6}  & {38.8} & {50.0}\\
{LinGlo-b1(Ours)} & {23.8}  & \textbf{41.5} & \textbf{62.4} & \textbf{44.5} & \textbf{27.1}  & \textbf{45.4} & \textbf{53.9}\\
\noalign{\smallskip}
\hline
\noalign{\smallskip}

{ResNet50}\cite{resnet} & {37.7}  & {36.3} & {55.3} & {38.6} & {19.3}  & {40.0} & {48.8}\\
{PVT-Small}\cite{PVT} & {34.2}  & {40.4} & {61.3} & {43.0} & {25.0}  & {42.9} & {55.7}\\
{Twins-PCPVT-S}\cite{Twins} & {34.4}  & {43.0} & {64.1} & {46.0} & {27.5}  & {46.3} & {57.3}\\
{Twins-SVT-S}\cite{Twins} & {34.3}  & {42.3} & {63.4} & {45.2} & {26.0}  & {45.5} & {56.5}\\
{Swin-T}\cite{Swin} & {38.5}  & {41.5} & {62.1} & {44.2} & {25.1}  & {44.9} & {55.5}\\
{LinGlo-b2(Ours)} & {35.1} & \textbf{44.7} & \textbf{65.8} & \textbf{47.9} & \textbf{28.0} & \textbf{48.9} & \textbf{57.8} \\
\hline
\end{tabular}
\end{center}
\end{table}
\setlength{\tabcolsep}{1.4pt}
\subsection{Semantic Segmentation}
\textbf{Setting.} ADE20K\cite{ade20k} contains 150 fine-grained semantic categories, with 20210, 2000, and 3352 images for training, validation, and testing, respectively. Following PVT\cite{PVT}, we utilize SemanticFPN\cite{semanticfpn} as our base framework.We use the weights pre-trained on ImageNet-1K to initialize the backbone and Xavier\cite{Xavier} to initialize the newly added layers. All models are optimized by AdamW\cite{adamw} with an initial learning rate of 0.0001.The learning rate is decayed following the polynomial decay schedule with a power of 0.9. For fair comparision, we use a batch size of 32 and train models for 40k iterations as PVT actually does according to their published code. We randomly resize and crop the image to 512$\times $512 for training, and rescale to have a shorter side of 512 pixels during testing. 

\textbf{Result.} As shown in Table~\ref{table:ade20k_result}, LinGlos consistently outperforms the counterparts. With similar parameters, LinGlo-b1 is 6.8 mIoU higher than PVT-Tiny, and LinGlo-b2 is 5.4 mIoU higher than PVT-Small. Compare to Swin transformer-Tiny, LinGlo-b2 is 3.7 mIoU higher with fewer parameters.
\setlength{\tabcolsep}{4pt}
\begin{table}
\begin{center}
\caption{Semantic segmentation result on ADE20K dataset.SemanticFPN is the framework.LinGlos consistently outperforms the counterparts with similar parameters.}
\label{table:ade20k_result}
\begin{tabular}{lcc}
\hline\noalign{\smallskip}
Backbone & Param(M) & mIoU(\%)\\
\noalign{\smallskip}
\hline
\noalign{\smallskip}
LinGlo-b0(Ours)  & {7.6} & {37.3} \\
\noalign{\smallskip}
\hline
\noalign{\smallskip}
{ResNet18} & {15.5}  & {32.9} \\
{PVT-Tiny} & {17.0}  & {35.7} \\
{LinGlo-b1(Ours)} & {17.8}  & \textbf{42.5} \\
\noalign{\smallskip}
\hline
\noalign{\smallskip}
{ResNet50} & {28.5}  & {36.7} \\
{PVT-Small} & {28.2}  & {39.8} \\
{Twins-PCPVT-S} & {28.4}  & {44.3} \\
{Twins-SVT-S} & {28.3}  & {42.6}\\
{Swin-T} & {31.9}  & {41.5}\\
{LinGlo-b2(Ours)} & {29.1} & \textbf{45.2} \\
\hline
\end{tabular}
\end{center}
\end{table}
\setlength{\tabcolsep}{1.4pt}

\section{Ablation Study}
In this section, we ablate important design elements in the proposed LinGlos, using ImageNet-1K image classification.
The experimental settings are the same as the settings in Sec.4.1 except that all models are trained for 100 epochs and 3 epochs for warming up.
\subsection{Vertical Layout Design}
To verify the effectiveness of alternate layout, we gradually replace the encoder layers in stage 1 and stage 2 of LinGlo-b0 by convolutional blocks with SE\cite{SENet}, which leads to two grafted architectures: C-H-H-H and C-C-H-H, where C denotes convolution and H denotes hybridization of convolution and attention.The parameter numbers of C-H-H-H and C-C-H-H are are almost the same as that of LinGlo-b0.The result in Table~\ref{table:grafting} shows that LinGlo-b0 \textgreater C-H-H-H \textgreater C-C-H-H, implying that the more restricted the use of spatial attention in early stages, the worse performance. LinGlo-b0 performs best. This may be because all stages in LinGlo-b0 can benefit from both spatial attentions and convolutions.
\setlength{\tabcolsep}{4pt}
\begin{table}
\begin{center}
\caption{Gradually replacing the encoder layers by convolutional blocks with SE .The more restricted the use of spatial attention in early stages, the worse performance.}
\label{table:grafting}
\begin{tabular}{lcc}
\hline\noalign{\smallskip}
Method & Params(M) & Top-1 Acc(\%)\\
\noalign{\smallskip}
\hline
\noalign{\smallskip}
C-C-H-H   & 3.3M & {67.2} \\
C-H-H-H   & 3.4M & {68.8} \\
H-H-H-H (i.e. LinGlo-b0) &  3.4M & \textbf{69.7} \\
\hline
\end{tabular}
\end{center}
\end{table}
\setlength{\tabcolsep}{1.4pt}

\subsection{Key-Only Attention}
\begin{figure}
\centering
\includegraphics[height=5cm]{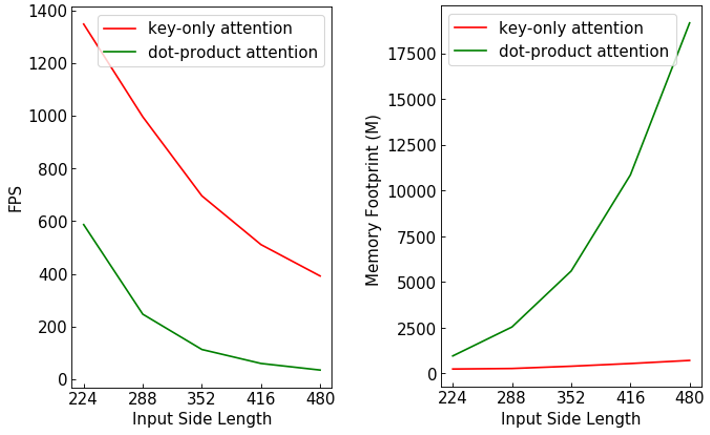}
\caption{Memory footprints and latency measured on 32GB V100 GPU. Batch size is 16.Key-only attention is much more memory and compute efficient}
\label{fig:mem}
\end{figure}

We replace all key-only attention in LinGlo-b0 with dot-product attention, and compare memory footprints and latency with various input sizes and a batch size of 16.This is a direct comparison between key-only attention and dot-product attention on a fixed architecture.The experiment is conducted on a 32GB V100 GPU.As shown in Fig.~\ref{fig:mem},key-only attention is much faster than dot-product attention.
As the input size increases, Key-only attention's memory footprint increases very slowly.This suggests that key-only attention may be well suited for use in
memory constrained scenarios.

\subsection{RP-CPE}
We ablate each component of RP-CPE to verify its effectiveness.By removing skip-connection ,we obtain parallel conditional positional encoding, denoted as P-CPE. By removing the other parallel CPE branches and keeping the skip-connection, we obtain residual conditional positional encoding, denoted as R-CPE. The comparision result is in Table~\ref{table:rp-cpe}.Adding a skip connection and combining multiple CPE branches both yield improvement. Because all parallel branches can be re-parameterized into a single branch, the improvement is cost-free. Adding more parallel branches or diverse branches could potentially yield further improvement\cite{DBB,repvgg,acnet},  but is outside the scope of this work .

\setlength{\tabcolsep}{4pt}
\begin{table}
\begin{center}
\caption{Ablating each component of RP-CPE.Adding a skip connection and combining multiple CPE branches in parallel both yield further improvement.}
\label{table:rp-cpe}
\begin{tabular}{lc}
\hline\noalign{\smallskip}
Method & Top-1 Acc(\%)\\
\noalign{\smallskip}
\hline
\noalign{\smallskip}
CPE   & {68.5} \\
R-CPE   & {69.2} \\
P-CPE   & {69.3} \\
RP-CPE   & \textbf{69.7} \\
\hline
\end{tabular}
\end{center}
\end{table}
\setlength{\tabcolsep}{1.4pt}
\subsection{Inductive Biases}
We ablate some inductive biases used in designing LinGlos, including hierarchical representation, and using convolutions for translation equivariance. 
By replacing convolutions in down-sampling layers with non-overlap patch embeddings\cite{Swin}, and removing all RP-CPEs, we obtained an architecture  denoted as LinGlo-w/o-C. The fist down-sampling layer of LinGlo-w/o-C is 4$\times$4 patch embedding, and others are 2$\times$2 patch embeddings.LinGlo-w/o-C uses absolute position encoding as in\cite{PVT} and has hierarchical structure.

Further, we ablate the hierarchical representation  by using 16$\times$16 non-overlap \cite{ViT} patch embedding in LinGlo-w/o-C's first down-sampling layer and removing all other down-sampling layers. The obtained architecture is denoted as LinGlo-w/o-CH. The channel number is set to 150 in all stages so that parameters are similar to LinGlo-b0's. The attention head number is set to 3 as in \cite{DeiT}.

\begin{figure}
\centering
\includegraphics[height=4cm]{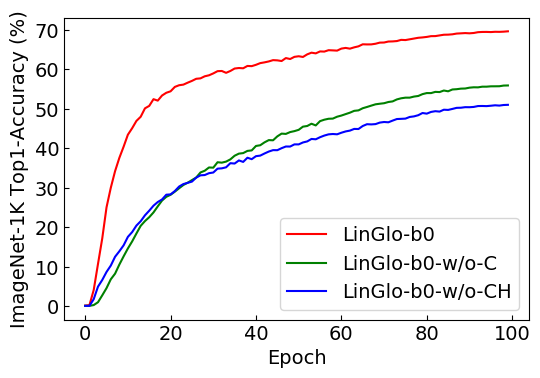}
\caption{Ablating inductive biases.LinGlo-b0 outperforms LinGlo-w/o-C(ablating convolutions) and LinGlo-w/o-CH(ablating convolutions and hierarchical representation)}
\label{fig:pvtele_deitele}
\end{figure}

In Fig.~\ref{fig:pvtele_deitele}, we plot the accuracy curves of LinGlo-b0, LinGlo-w/o-C-b0 and LinGlo-w/o-CH. It shows that integrating these inductive biases can not only accelerate convergence, but also improve peak performance significantly.

\section{Conclusions and Future Work}
We introduce a new self-attention visual backbone family, LinGlos.To address the issue of attention's computational inefficiency on high-resolution inputs, we use key-only attention to establish global-dependency in all stages without pairwise interactions. Experiments on image classification,object detection and semantic segmentation benchmarks verify that our LinGlos are stronger than CNNs , pure transformers and hybrid backbones under similar parameters.

This paper currently focuses on parameter-limited setting. However, we believe our approach is applicable to large models. Besides, since LinGlos' computational complexity is linear w.r.t. spatial size, it's very promising to apply our approach to visual tasks with high-resolution inputs or outputs, such as super-resolution. We will leave them for future work.

\clearpage
%
%

\end{document}